\documentclass[10pt, a4paper, twocolumn]{article} 

\usepackage[english]{babel} 

\usepackage{microtype} 

\usepackage{amsmath,amsfonts,amsthm} 

\usepackage[svgnames]{xcolor} 

\usepackage[hang, small, labelfont=bf, up, textfont=it]{caption} 

\usepackage{booktabs} 

\usepackage{lastpage} 

\usepackage{graphicx} 

\usepackage{enumitem} 
\setlist{noitemsep} 

\usepackage{sectsty} 
\allsectionsfont{\usefont{OT1}{phv}{b}{n}} 

\usepackage{geometry} 

\usepackage[T1]{fontenc} 
\usepackage[utf8]{inputenc} 

\usepackage{XCharter} 

\usepackage{fancyhdr} 
\fancypagestyle{firstpage}{ 
	\fancyhf{}
}

\newcommand{\authorstyle}[1]{{\large\usefont{OT1}{phv}{b}{n}\color{DarkRed}#1}} 

\newcommand{\institution}[1]{{\footnotesize\usefont{OT1}{phv}{m}{sl}\color{Black}#1}} 

\usepackage{titling} 

\newcommand{\HorRule}{\color{DarkGoldenrod}\rule{\linewidth}{1pt}} 

\pretitle{
	\vspace{-30pt} 
	\HorRule\vspace{10pt} 
	\fontsize{32}{34}\usefont{OT1}{phv}{b}{n}\selectfont 
	\color{DarkRed} 
}

\posttitle{\par\vskip 15pt} 

\preauthor{} 

\postauthor{ 
	\vspace{10pt} 
	\par\HorRule 
	\vspace{20pt} 
}

\usepackage{lettrine} 
\usepackage{fix-cm}	

\newcommand{\initial}[1]{ 
	\lettrine[lines=3,findent=4pt,nindent=0pt]{
		\color{DarkGoldenrod}
		{#1}
	}{}%
}

\usepackage{xstring} 

\newcommand{\lettrineabstract}[1]{
	\StrLeft{#1}{1}[\firstletter] 
	\initial{\firstletter}\textbf{\StrGobbleLeft{#1}{1}} 
}

\usepackage{times}
\usepackage{epsfig}
\usepackage{graphicx}
\usepackage{amsmath}
\usepackage{amssymb}
\usepackage{microtype}
\usepackage[pagebackref=true,breaklinks=true,letterpaper=true,colorlinks,bookmarks=false]{hyperref}

\title{A Unified Multi-Faceted Video Summarization System}

\author{
	\authorstyle{Anurag Sahoo\textsuperscript{1}, Vishal Kaushal\textsuperscript{1,2}, Khoshrav Doctor\textsuperscript{1}, Suyash Shetty\textsuperscript{1}, Rishabh Iyer\textsuperscript{3} and Ganesh Ramakrishnan\textsuperscript{2}} 
	\newline\newline 
	\textsuperscript{1}\institution{AITOE LABS}\\ 
	\textsuperscript{2}\institution{Indian Institute of Technology, Bombay}\\ 
	\textsuperscript{3}\institution{Microsoft Corporation} 
}

\date{\today} 

\begin{document}
\maketitle
\thispagestyle{firstpage} 
\lettrineabstract{This paper addresses automatic summarization and search in visual data comprising of videos, live streams and image collections in a unified manner. In particular, we propose a framework for multi-faceted summarization which extracts key-frames (image summaries), skims (video summaries) and entity summaries (summarization at the level of entities like objects, scenes, humans and faces in the video). The user can either view these as extractive summarization, or query focused summarization. Our approach first pre-processes the video or image collection once, to extract all important visual features, following which we provide an interactive mechanism to the user to summarize the video based on their choice. We investigate several diversity, coverage and representation models for all these problems, and argue the utility of these different models depending on the application. While most of the prior work on submodular summarization approaches has focused on combining several models and learning weighted mixtures, we focus on the explain-ability of different the diversity, coverage and representation models and their scalability. Most importantly, we also show that we can summarize hours of video data in a few seconds, and our system allows the user to generate summaries of various lengths and types interactively on the fly.}

\section{Introduction}
Visual Data in the form of images, videos and live streams have been growing at an unprecedented rate in the last few years. While this massive data is a blessing to data science by helping improve predictive accuracy, it is also a curse since humans are unable to consume this large amount of data. Moreover, today, machine generated videos (via Drones, Dash-cams, Body-cams, Security cameras, Go-pro etc.) are being generated at a rate higher than what we as humans can process. Moreover, majority of this data is plagued with redundancy. Given this data explosion, machine learning techniques which automatically understand, organize and categorize this data are of utmost importance. This paper attempts to provide a framework and a system for Visual Data Organization, Categorization and Summarization. This work presents a \emph{Summarization System} which provides a quick snapshot and summary of the video (equivalently collection of photographs) while saving the user time to understand the video. The goal of the system is to interact with the user to quickly provide relevant content.

The input to our system is a video or an image collection. The video (or image collection) is analyzed and preprocessed by the system to produce an analysis database, which contains information of all the important visual concepts from the video, including the scenes, the objects, the humans, faces and colors and possibly other meta information. These features form the inputs to our system, along with other information such as the desired size of the summary video.  The following are the variants of summarization addressed through our framework:
\begin{itemize}
\item \emph{Extractive Summarization:} Extract a set of diverse, yet representative snippets or frames from the original video or image collection. The summary can be determined at runtime based on the desired length of the summary, or the desired coverage, representation and diversity. This summary can be in the form of a video or a set of images.
\item \emph{Query Focused Summarization or Search: } Given a query, the task here is to extract a representative set of frames or snippets relevant to that query. The system quickly retrieves the relevant frames or snippets at runtime (details described in the rest of the paper).  Once the related keyframes or snippets are extracted, the procedure is similar to extractive summarization. The output is a video skim or set of keyframes relevant to this query.
\end{itemize}
In both the above variants, we provide a multi-faceted video summary in the form of
\begin{enumerate}
\item a video skim, i.e. a video summary containing a subset of the frames as a continuous video
\item an image summary in the form of key-frames from the video
\item Entity based summarization (which we also call Concept based summarization)
\end{enumerate}
Entity based summarization goes a step beyond skims, keyframes and focuses on entities, like objects, scenes, humans, faces to provide a representative yet diverse subset of these entities. This answers questions like who are the different people or what are the diverse objects and scenes in the video. Finally, note that since videos consist of of a set of frames, the problem of extracting keyframes from an input video generalizes the problem of image collection summarization~\cite{tschiatschek2014learning,simon2007scene}, hence without loss of generality, we shall henceforth focus on \emph{Video Summarization}. Towards the end of this paper, we discuss how we can handle summarization of live video streams.

\subsection{Existing Work}
Several papers in the past have investigated the problems of video and image collection summarization. Video Summarization techniques differ in the way they generate the output summary. Some of these~\cite{wolf1996key,lee2012discovering,khosla2013large,kim2014joint} extract a set of keyframes from the video, while others focus on extracting video summaries or skims from the long video~\cite{gygli2015video,gygli2014creating,zhang2016summary,zhao2014quasi}. Other forms of video summarization include creating GIF summaries from videos~\cite{gygli2016video2gif}, Montages~\cite{sun2014salient}, Visual Storyboards from videos~\cite{goldman2006schematic,lu2013story}, video synopses~\cite{pritch2008nonchronological} and time lapses and hyperlapse summaries~\cite{kopf2014first}. Similarly, image collection summarization involves choosing a subset of \emph{representative}  images from the collection~\cite{tschiatschek2014learning,simon2007scene}. Another line of approach, which is similar to what we call \emph{Entity based Summarization}, was proposed in \cite{meng2016keyframes}, wherein the authors select representative summaries of all objects in a video. They do this by modeling the problem as that of sparse dictionary selection.

Most video summarization techniques can be categorized into methods trying to model one of three properties of summaries (i) interestingness (how good is a given snippet as a summary), (ii) representativeness (how well the summary represents the entire video or image collection), and (iii) diversity (how non-redundant and diverse is the summary). Examples of methods which model interestingness of snippets include~\cite{wolf1996key} that find summary snippets through motion analysis and optical flow, \cite{lee2012discovering} which uses humans and objects to determine interesting snippets and finally, \cite{gygli2014creating} which models interestingness through a \emph{superframe} segmentation. \cite{chu2015video} summarizes multiple videos collectively by looking at inter-video-frame similarity and posing a maximal bi clique finding algorithm for finding summaries. Methods which only model the quality of the snippets, or equivalently the interestingness of the summaries and do not model the diversity often achieve redundant frames and snippets within their summary.

Hence a lot of recent work has focused on diversity models for video and image collection summarization. \cite{simon2007scene} used the Facility Location function with a diversity penalty for image collection summarization, while \cite{sinha2011extractive} defined a coverage function and a disparity function as a diversity model. \cite{lu2013story} attempted to find the candidate chain of sub shots that has the maximum score composed of measures of story progress between sub shots, importance of individual sub shots and diversity among sub shot transitions.   \cite{tschiatschek2014learning} was among the first to use a mixture of submodular functions learnt via human image summaries for this problem. For video summarization, \cite{li2010multi} proposed the Maximum Marginal Relevance (MMR) as a diversity model, while \cite{zhang2016summary, gong2014diverse} used a Determinantal Point Process based approach for selecting diverse summaries. \cite{zhao2014quasi} proposed an approach for video summarization based on dictionary based sparse coding, and \cite{gygli2015video} proposed using mixtures of submodular functions and supervised learning of these mixtures via max-margin training, an approach used for several other tasks including document summarization~\cite{lin2012learning} and image collection summarization~\cite{tschiatschek2014learning}.

\subsection{Our Contributions}
As observed in prior work~\cite{gygli2015video,lin2012learning, lin2011class,tschiatschek2014learning} several models for diversity, representation, coverage and uniformity can be unified within the class of Submodular Optimization. This includes, for example, Determinantal Point Processes, Maximum Marginal Relevance, Facility Location, Disparity Functions and several others. This paper attempts to unify these formulations, and provide a complete summarization system in a multi-faceted way.

\begin{enumerate}
\item We present a unified approach for several variants of Video and Image collection summarization, including generating keyframes (summary comprised of image frames), skims (video summaries), and entity summaries (summary of all objects, scenes, people and faces in the video). We do this both for extractive and query focused versions. While earlier work has done this for a subset of these problems, we show how a unified submodular formulation works across all these variations.
\item We investigate several diversity, coverage and representation functions, and demonstrate how different models are applicable in different kinds of video summarization tasks. We validate our claims by empirically showing the behavior of these functions on different kinds of videos.
\item We discuss the computational scalability of the optimization algorithms, and point out some computational tricks including lazy evaluations and memoization, which enable optimized implementations for various submodular functions. As a result, we show that once the important visual features have been extracted (via a pre-processing step), we can obtain the summary subset of the video (or frames) in a few seconds. This allows the user to interactively obtain summaries of various lengths, types and queries in real time.
\item We show several experiments of our system demonstrating the computational scalability and modeling capabilities.
\end{enumerate}

Most past work on Video and Image collection summarization, either use a subset of hand-tuned submodular functions~\cite{simon2007scene,lin2011class, li2010multi} or a learnt mixture of submodular functions~\cite{gygli2015video,tschiatschek2014learning,lin2012learning}. This work addresses the orthagonal aspect of how do we build a unified submodular summarization system, how do different subclasses of submodular functions model summarization and what are the practical computational and algorithmic tricks in implementing such a system. We investigate several variants of video summarization and show how each of these problems can be unified within our submodular summarization system.
\section{Background and Main Ideas}
This section describes the building blocks of our framework, namely the Submodular Summarization Framework and the basics of Convolutional Neural Networks for Image recognitions (to extract all the objects, scenes, faces, humans etc.)
\subsection{Submodular Summarization Framework}
We assume we are given a set $V = \{1, 2, 3, \cdots, n\}$ of items which we also call the \emph{Ground Set}. Also define a utility function $f:2^V \rightarrow \mathbf{R}$, which measures how good of a summary a set $X \subseteq V$ is. Let $c :2^V \rightarrow \mathbf{R}$ be a cost function, which describes the cost of the set (for example, the size of the subset). The goal is then to have a summary set $X$ which maximizes $f$ while simultaneously minimizes the cost function $c$. In this paper, we study a special class of set functions called \emph{Submodular Functions}. Given two subsets $X \subseteq Y \subseteq V$, a set function $f$ is submodular, if $f(X \cup j) - f(X) \geq f(Y \cup j) - f(j)$, for $j \notin Y$. This is also called the diminishing returns property.  Several Diversity and Coverage functions are submodular, since they satisfy this diminishing returns property. We also call a function \emph{Monotone Submodular} if $f(X) \leq f(Y)$, if $X \subseteq Y \subseteq V$. The ground-set $V$ and the items $\{1, 2, \cdots, n\}$ depend on the choice of the task at hand. We now define a few relevant optimization problems which shall come up in our problem formulations:

\begin{align}
\mbox{Problem 1:} \max_{X \subseteq V, |X| = k} f(X)
\end{align}
Problem 1 is cardinality constrained submodular maximization~\cite{nemhauser1978analysis}, and $k$ is the cardinality (size) constraint on the summary. This a natural model for extracting fixed length summary videos (or a fixed number of keyframes).
\begin{align}
\mbox{Problem 2:} \max_{X \subseteq V, S(X) \leq b} f(X)
\end{align}
This is \emph{Knapsack Constrained Submodular Maximization}~\cite{sviridenko2004note}. The goal here is to find a summary with a fixed cost, and $S_1, S_2, \cdots, S_n$ denotes the cose of each element in the ground-set.
\begin{align}
\mbox{Problem 3: } \min_{f(X) \geq c} S(X)
\end{align}
This problem is called the \emph{Submodular Cover Problem}~\cite{wolsey1982analysis, iyer2013submodular}. $S(X)$ is the cost function which is modular, and  $c$ is the coverage constraint. The goal here is to find a minimum cost subset $X$ such that the submodular coverage or representation function covers \emph{information} from the ground set. A special case of this is the set cover problem. Moreover, Problem 3 can be seen as a Dual version of Problem 2~\cite{iyer2013submodular}.

Submodular Functions have been used for several summarization tasks including Image summarization~\cite{tschiatschek2014learning}, video summarization~\cite{gygli2015video}, document summarization~\cite{lin2011class}, training data summarization and active learning~\cite{wei2015submodularity} etc. Using a
greedy algorithm to optimize a submodular function (for selecting a subset) gives a lower-bound performance guarantee of around 63\% of optimal and in practice these greedy solutions are often within 90\% of optimal ~\cite{krause2008optimizing}. This makes it advantageous to formulate (or approximate) the objective function for data selection as a submodular function.

\subsection{CNNs for Image Feature Extraction}
Convolutional Neural Networks are critical to feature extraction in our summarization framework. We pre-process the video to extract key visual features including objects, scenes, faces, humans, faces, texts and actions. Convolutional Neural Networks have recently provided state of the art results for several recognition tasks including object recognition~\cite{krizhevsky2012imagenet, szegedy2015going, he2016deep}, Scene recognition~\cite{zhou2014learning}, Face Recognition~\cite{parkhi2015deep} and Object Detection and Localization~\cite{redmon2016you}. We next describe the end to end system in detail.

\section{Method}
The input to our system is a video or an image collection. Our system then extracts all important features from the video and generates an analysis database. The user can then interact with the system in several ways. User can generate a video summary of a given length, or extract a set of key frames or a montage describing the video. Similarly the user can search for a query and extract relevant video snippets of frames which are relevant to the query. Finally the user can also view a summary of all objects, scenes, humans and faces in the video along with their statistics. All these interactions are enabled on the fly (in a few seconds). The user can also define the summarization model of their choice. We investigate and compare  different submodular models, and argue the utility of different models based on the use case.
\subsection{Problem Formulation for the Multi-Faceted Visual Summarization}
We now formulate problem statements across different summarization views on videos or image collections.
\subsubsection{Extractive Video Summarization}
Extractive Video Summaries can be provided as Video Keyframes (a collection of frames), a video skim or a summary based on entities in the video.
\begin{figure}
\includegraphics[width=\linewidth]{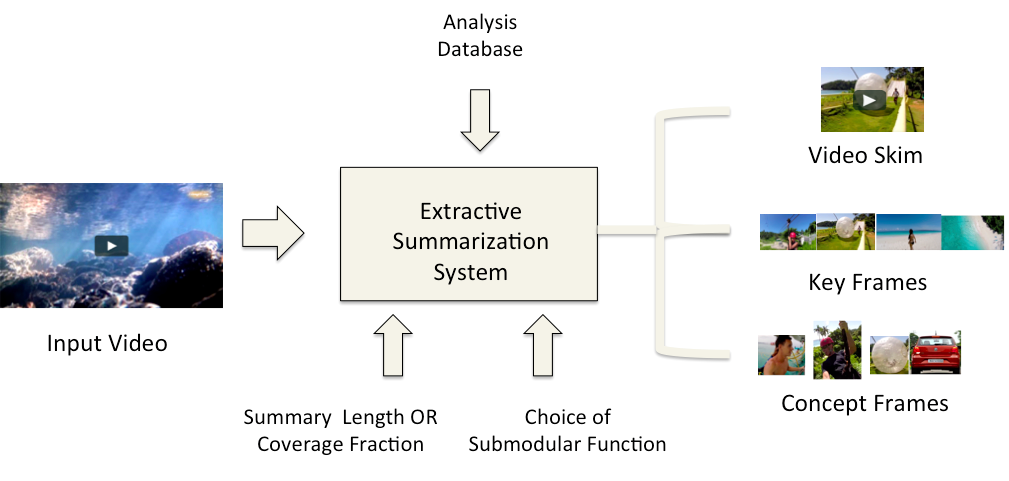}
\caption{Illustration of Extractive Video Summarization in the form of Video Skims, Key Frames and Entity Frames}
\label{figex}
\end{figure}
\paragraph{Video Summary as Key Frames}
The problem here is to generate a set of frames, called \emph{keyframes} which represent the video. The approach is to sample frames from a video at a given frame-rate (say, 1 frame per second) and use these as the ground-set $V$. Suppose we sample at a frame rate $F$ and the length of the video is $T$ seconds, the ground set size $|V| = FT$. In the case of Image collection summarization, the individual items are the images. The approach is to then find a subset of frames/images $X$ of size $k$ (number of key frames) which maximizes a submodular diversity or coverage function $f$. This becomes an instance of Problem 1. Given a submodular coverage function, one can also ask what is the minimum size subset which \emph{covers} information from the entire video -- for example, a subset of frames which covers all the diverse scenes in a video or image collection. In that case, $f$ can be a coverage function, like a set cover, and the goal is to find a minimum size set of frames, $X$ such that $f(X) \geq c = f(V)$, where again, $V$ is the ground-set of frames.
\paragraph{Video Summary as Skims}
The task is to generate a video skim (summary) of the longer video. We first consider a fixed budget summarization problem, i.e. generate a video summary of a desired length. There are several ways we can represent the individual items and the ground set, as listed below.
\begin{enumerate}
\item The simplest choice of the items here are fixed length snippets $S$. For example, each element in the ground set is a two second snippet of the video. In this setting, the cost function is simply the cardinality, since the length of the summary video is directly proportional to the size of the summary subset. In particular, denote $T$ as the length of the entire video and $B$ as the budget (in seconds) of the summary. Then, $V$ consists of snippets of size $S$ and $|V| = T/S$, and $k = B/S$.  This becomes an instance of Problem 1.
\item Another choice is to have individual shots as the snippets. The advantage of that is it ensures the summary video is not abruptly cut visually. Unlike the first case, each snippet does not have the same length (in seconds). In particular, assume the video has $s$ shots, and let $S_1, S_2, \cdots, S_s$ denote the length of each snippet (shot). This then becomes an instance of Problem 2, where we want to maximize $f$ such that $S(X) \leq B$ where $B$ is the budget (in seconds) of the summary video.
\item Yet another choice is defining a snippet based on meta data like subtitles and the speech. A simple way is to detect speech boundaries, or if subtitles are available, use the subtitles to determine the boundaries. This ensures that the audio and speech is not abruptly cut in the summaries. Again, here, the snippets are not of the same length and problem is very similar to the above.
\end{enumerate}
In each of the cases above, one can also generate a video summary which retains all \emph{information} within a video while removing all redundant snippets from the video. This then becomes an instance of Problem 3, where we want to minimize $S(X)$ or the length of the summary such that $f(X) = f(V)$.
\paragraph{Entity based Summarization}
The task is to generate a summary of the  \emph{objects, faces, scenes} and \emph{humans} extracted in the video. Again, while preprocessing the video, we extract all the objects, scene images, faces and humans from the video. Denote, for example, $V_o$ as the set of extracted objects on analyzing each frame of the video. Similarly denote $V_h$ as humans and $V_f$ as faces. Note that many of these will be comprised of redundant images. The goal is to find a fixed size summary subset which is diverse and representative (Problem 1) or a subset retaining all information of the objects/humans/faces while removing redundancy (Problem 3). This is useful in generating statistics of objects/scenes etc. and giving a visual representation of these in the video.

Figure~\ref{figex} illustrates the three kinds of extractive summarization.
\begin{figure}
\includegraphics[width=\linewidth]{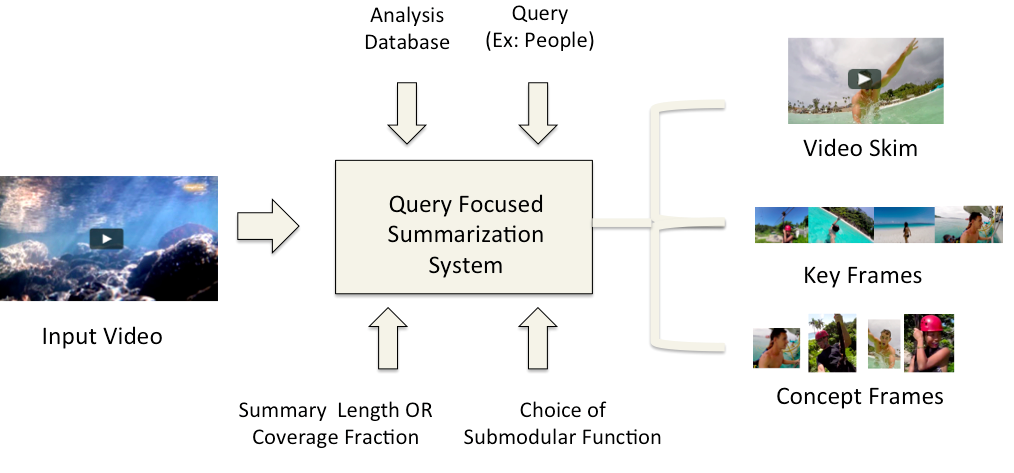}
\caption{Illustration of Query Focussed Video Summarization in the form of Video Skims, Key Frames and Entity Frames. In each case, we extract the summary relevant to the query (ex: people in this case)}
\label{figq}
\end{figure}
\subsubsection{Query Based Summarization}
Query Based Summarization also generates keyframes (image summary), video skims or concept frames as summary. Unlike extractive summarization, query summarization just focuses on video frames (in the case of key frame summary), video snippets (in the case of video skims) or entities (objects, scenes, faces etc.) which are relevant to the query. Given a query $q$, we first extract a set $V_q$ of snippets, frames or concepts relevant to the query. This is done by querying the analysis database generated while preprocessing the video. The generation of the analysis database is described in the next subsection. In particular, the analysis database contains labels of objects, scenes, humans with their age and gender, along with color information. Once we extract a subset of relevant query snippets of frames, we then follow the procedure similar to that of extractive summarization, depending on whether the user is interested in generating key-frames or video skims. Figure~\ref{figq} illustrates query based summarization.
\begin{figure}
\includegraphics[width=\linewidth]{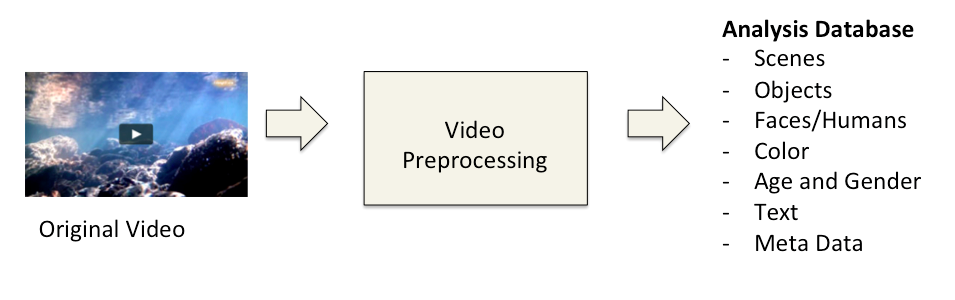}
\caption{Illustration of the Video Preprocessing steps to generate the analysis database.}
\end{figure}
\subsection{Video Preprocessing}
This section describes in detail the preprocessing steps to generate key features used in Summarization, Search and generating Statistics in a Video. All features are extracted at a fixed frame rate (we use one frame per second) and stored.
\subsubsection{Scene Features}
We use a CNN proposed in~\cite{zhou2014learning} trained on Places205 Dataset. We extract the a) scene labels, b) Final layer probabilities and c) the features. The Features are extracted from the second last layer of the CNN, and the probabilities are extracted after the soft-max layer. We also extract the labels based on the output of the CNN, when the prediction probability exceeds a given threshold. We do this at the given fixed frame rate. The labels, probabilities as well as the features are stored in the analysis file.
\subsubsection{Object Features}
For object detection, we use YOLO~\cite{redmon2016you} to extract localized object labels, along with features extracted and the probabilities for each object. YOLO (also called DarkNet) localizes objects in the frame, so store the set of objects detected (the labels), along with their position, size, feature representation and probability. Note that this is done for every object localized, and not at a frame level. We also extract features at the frame-level, where we use GoogLeNet~\cite{krizhevsky2012imagenet, szegedy2015going} model, pretrained on ImageNet. We store the labels, final layer probabilities and the features from the second last layer of the CNN.
The localized and frame-level features are extracted at the fixed framerate.
\subsubsection{Face Features}
In order to extract Facial features, we first detect a Face. For Face detection, we use the Histogram of Gradients~\cite{dalal2005histograms} trained on Faces. We also store the position and size of each detected face, along with the features extracted using the VGG Face Model~\cite{parkhi2015deep}.
\subsubsection{Human Features}
For Human detection, we use two kinds of algorithms. One is a full body (pedestrian detection), where we use Histogram of Gradients based human detectors~\cite{dalal2005histograms} and Deformable parts model~\cite{felzenszwalb2008discriminatively}, while we use YOLO~\cite{redmon2016you} for detecting other people (in cases when the full body is not visible). We found that the full body detectors work well when the full body of the people are visible, while YOLO works well when the people are closer to the camera. We store the position and size of each detected human, and we extract features for each human using GoogLeNet~\cite{szegedy2015going}.
\subsubsection{Age and Gender Features}
For each detected Face and Human, we identify the Age and Gender, following the architecture in~\cite{levi2015age}. We also extract features from this model for each detected Face and Human, along with the labels and probabilities.
\subsubsection{Color Features}
For color features, we store the color histogram of each frame. We also store the color histogram corresponding to each detected object and human. We also recognize colors in objects and detected humans (shirt color) using techniques similar to~\cite{swain1991color}.

\begin{figure*}
\begin{center}
\includegraphics[width = 0.3\textwidth]{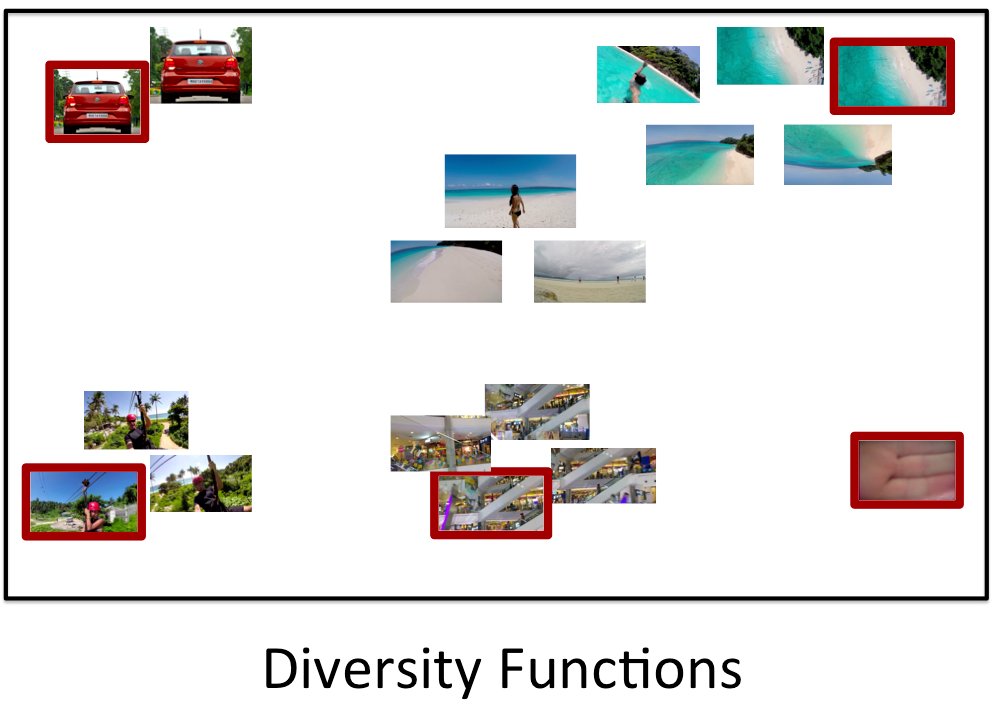}
\includegraphics[width = 0.3\textwidth]{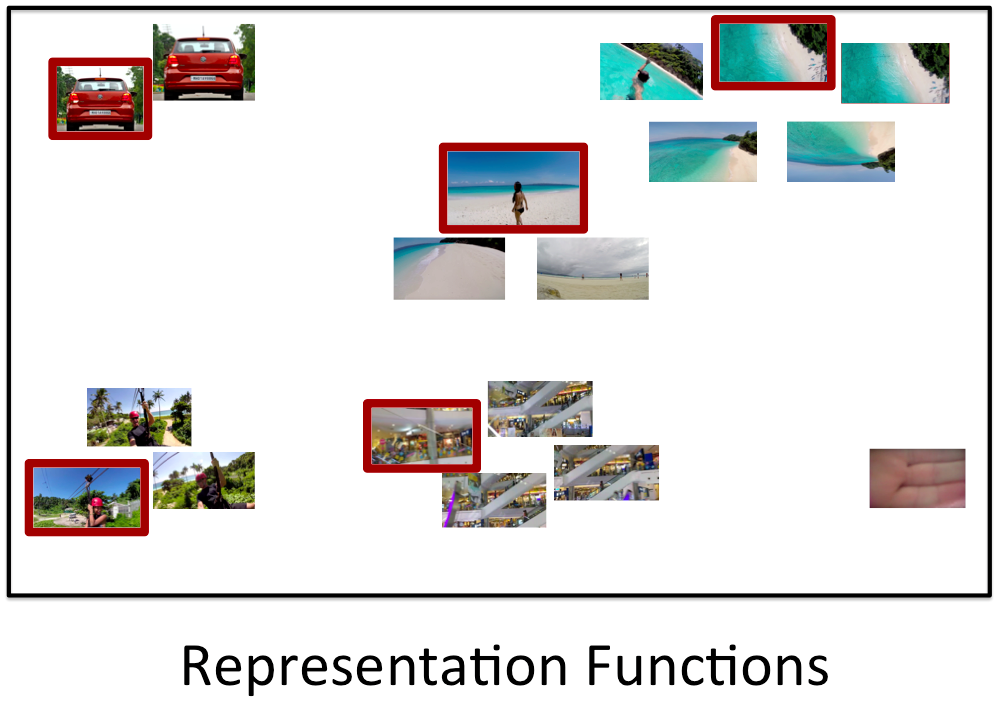}
\includegraphics[width = 0.3\textwidth]{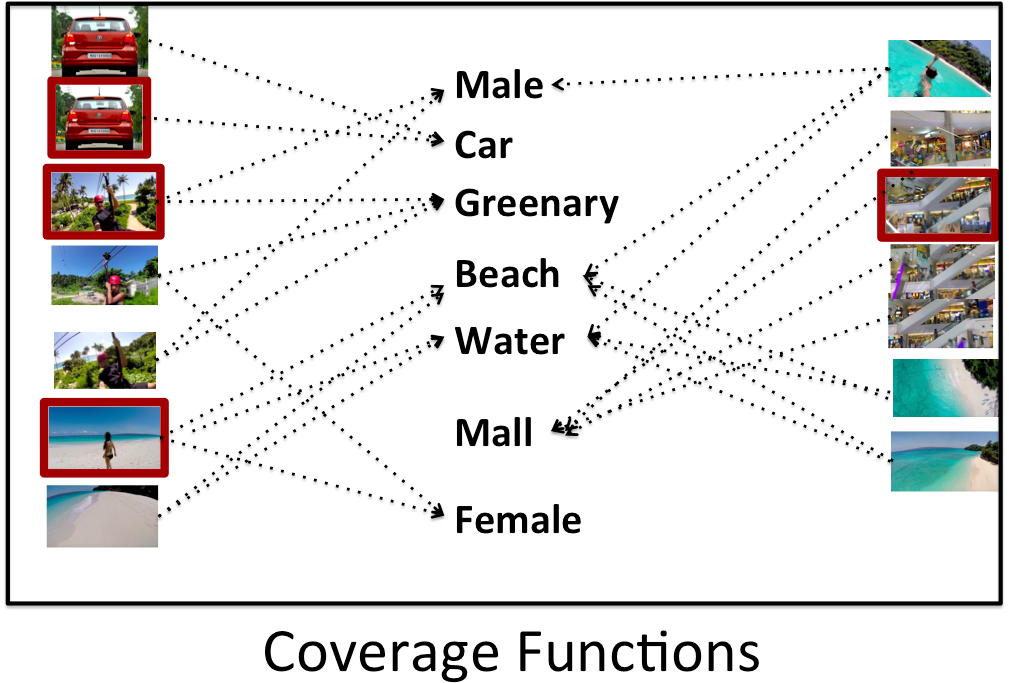}
\end{center}
\label{divrepcov}
\caption{Illustration of the Difference between Diversity Functions, Coverage Functions and Representation Functions}
\end{figure*}

\subsection{Submodular Functions for Summarization}
This section describes the Submodular Functions used in our system. We divide these into Coverage Functions, Representation Functions and Diversity Functions.
\subsubsection{Coverage Based Submodular Functions}
This class of functions model notions of coverage, i.e. try to find a subset of the ground set $X$ which covers a set of \emph{concepts}. Below are instantiations of this.
\paragraph{Set Cover Function:} Denote $V$ as the ground set and let $X \subseteq V$ be a subset (of snippets or frames). Further $\mathcal U$ denotes a set of concepts, which could represent, for example, scenes or objects. Each frame (or snippet) $i \in X$ contains a subset $U_i \in \mathcal U$ set of concepts (for example, an image covers a table, chair and person). The set cover function then is
\begin{align}
f(X) = w(\cup_{i \in X} U_i),
\end{align}
where $w_u$ denotes the weight of concept $u$.
\paragraph{Probabilistic Set Cover: } This is a generalization of the set cover function, to include probabilities $p_{i u_i}$ for each object $u_i$ in Image $i \in X$. For example, our convolutional neural network might output a confidence of object $u_i$ in Image $i$, and we can use that in our function. The probabilistic coverage function is defined as,
\begin{align}
f(X) = \sum_{i \in \mathcal U} w_i[1 - \prod_{i \in X}(1 - p_{ij})].
\end{align}
The set cover function is a special case of this if $p_{ij} = 1$ if Object $j$ belongs to Image $i$ (i.e. we use the hard labels instead of probabilities).
\paragraph{Feature Based Functions: } Finally we investigate the class of Feature Based functions. Here, we denote an Image $i$ via a feature representation $q_i$. This could be, for example, the features extracted from the second last layer of a ConvNet. Denote $F$ as the set of features. The feature based function is defined as,
\begin{align}
f(X) = \sum_{i \in F} \psi(q_i(X))
\end{align}
where $q_i(X) = \sum_{j \in X} q_{ij}$, and $q_{ij}$ is the value of feature $i$ in Image $j$. $\psi$ is a concave function.  Examples of $\psi$ are square-root, Log and Inverse Function etc.
\subsubsection{Representation based Submodular Functions}
Representation based functions attempt to directly model representation, in that they try to find a representative subset of items, akin to centroids and mediods in clustering.
\paragraph{Facility Location Function: } The Facility Location function is closely related to k-mediod clustering. Denote $s_{ij}$ as the similarity between images $i$ and $j$. We can then define $f(X) = \sum_{i \in V} \max_{j \in X} s_{ij}$. For each image $i$, we compute the representative from $X$ which is closest to $i$ and add the similarities for all images. Note that this function, requires computing a $O(n^2)$ similarity function. However, as shown in~\cite{wei2014fast}, we can approximate this with a nearest neighbor graph, which will require much smaller space requirement, and also can run much faster for large ground set sizes.
\subsubsection{Diversity Based Submodular Functions}
The third class of Functions are Diversity based ones, which attempt to obtain a diverse set of key points.
\paragraph{Disparity Function: } Denote $d_{ij}$ as a distance measure between Images $i$ and $j$. Define a set function $f(X) = \min_{i, j \in X} d_{ij}$. This function is not submodular, but can be efficiently optimized via a greedy algorithm~\cite{dasgupta2013summarization}. It is easy to see that maximizing this function involves obtaining a subset with maximal minimum pairwise distance, thereby ensuring a diverse subset of snippets or keyframes.

\paragraph{Determinantal Point Processes: } Another class of Functions are Determinantal Point Processes, defined as $p(X) = \mbox{Det}(S_X)$ where $S$ is a similarity kernel matrix, and $S_X$ denotes the rows and columns instantiated with elements in $X$. It turns out that $f(X) = \log p(X)$ is submodular, and hence can be efficiently optimized via the Greedy algorithm. Unlike the other choices of submodular functions investigated so far, this requires computing the determinant and is $O(n^3)$ where $n$ is the size of the ground set. This function is not computationally feasible and hence we do not use it in our system since we require near real time results in summarization.
\begin{table*}
\begin{center}
 \begin{tabular}{|| c | c |  c | c | c ||}
 \hline
 $f(X)$ & $p_f(X)$ & $f(j | X, p_f)$ & $C_o$ & $C_p$\\ [0.5ex]
 \hline\hline
 $\sum_{i \in V} \max_{k \in X} s_{ik}$ & $[\max_{k \in X} s_{ik}, i \in V]$ & $\sum_{i \in V} \max(p_f^i(X), s_{ij}) - p_f^i(X)$ & $O(n^2)$ & $O(n)$\\
 \hline
 $\sum_{i \in \mathcal F} \psi(w_i(X))$ & $[w_i(X), i \in \mathcal F]$ & $\sum_{i \in \mathcal F} [\psi(p_f^i(X) + w_i(j)) - \psi(p_f^i(X))]$ & $O(n|\mathcal F|)$ & $O(|\mathcal F|)$ \\
 \hline
 $w(\cup_{i \in X} U_i)$ & $\cup_{i \in X} U_i$ & $w(P_f \cup U_j) - w(P_f)$ & $O(n|U|$ & $|U|$\\
 \hline
 $\min_{k,l  \in X, k \neq l} d_{kl}$ & $\min_{k, l \in X, k \neq l} d_{kl}$ & $\min\{p_f(X), \min_{k \in X} d_{kj}\} - p_f(X)$ & $O(|X|^2)$ & $O(|X|)$\\
 \hline
 $\sum_{i \in \mathcal U} w_i[1 - \prod_{k \in X}(1 - p_{ik})]$ & $[\prod_{k \in X} (1 - p_{ik}), i \in \mathcal U]$ & $\sum_{i \in \mathcal U} [p_f^i(X) - p_f^i(X)p_{ij}]$ & $O(n|\mathcal U|)$ & $O(|\mathcal U|)$ \\ [1ex]
 \hline
\end{tabular}
\caption{List of Submodular Functions used, with the precompute statistics $p_f(X)$, gain evaluated using the precomputed statistics $p_f(X)$ and finally $C_o$ as the cost of evaluation the function without memoization and $C_p$ as the cost with memoization. It is easy to see that memoization saves an order of magnitude in computation.}
\end{center}
\end{table*}

\subsubsection{Diversity, Representation and Coverage Functions}
Figure 4 demonstrates the intuition of using diversity, representation and coverage functions. Diversity based functions attempt to find the most different set of images. The leftmost figure in Fig. 4 demonstrates this. It is easy to see that the five most diverse images are picked up by the diversity function (Disparity Min), and moreover, the summary also contains the image with a hand covering the camera (the image on the right hand side bottom), which is an outlier. The middle figure demonstrates the summary obtained via a representation function (like Facility Location). The summary does not include outliers, but rather contains one representative image from each cluster. The diversity function on the other hand, does not try to achieve representation from every cluster. The third figure demonstrates coverage functions. The summary obtained via a coverage function (like Set Cover or Feature based function), covers all the concepts contained in the images (Male, Car, Greenery, Beach etc.). From a modeling perspective, coverage and representation functions discussed in this paper are monotone submodular, while diversity functions are not in general monotone.

\subsection{Optimization Algorithms}
The previous sections describe the models used in our system. We now investigate optimization algorithms which solve Problems 1-3. Variants of a greedy algorithm provide near optimal solutions with approximation guarantees for Problems 1-3~\cite{wolsey1982analysis,nemhauser1978analysis,sviridenko2004note, lin2010multi}.
\paragraph{Cardinality Constrained Submodular Maximization: } For cardinality constrained submodular maximization (Problem 1), a simple greedy algorithm provides a near optimal solution~\cite{nemhauser1978analysis}. Starting with $X^0 = \emptyset$, we sequentially update $X^{t+1} = X^t \cup \mbox{argmax}_{j \in V \backslash X^t} f(j | X^t)$, where $f(j | X) = f(X \cup j) - f(X)$ is the gain of adding element $j$ to set $X$. We run this till $t = k$ and $|X^t| = k$, where $k$ is the budget constraint. It is easy to see that the complexity of the greedy algorithm is $O(nkT_f)$ where $T_f$ is the complexity of evaluating the gain $f(j | X)$.
\paragraph{Knapsack Constrained Submodular Maximization: } For the Knapsack constrained version (Problem 2), the greedy algorithm is a slight variant, where at every iteration, we sequentially update $X^{t+1} = X^t \cup \mbox{argmax}_{j \in V \backslash X^t} \frac{f(j | X^t)}{c(j)}$. This algorithm has near optimal guarantees~\cite{lin2010multi}. The complexity of this algorithm is very similar to the one above.
\paragraph{Submodular Cover Problem: } For the Submodular Cover Problem (Problem 3), we again resort to a greedy procedure~\cite{wolsey1982analysis} which is near optimal. In this case, the update is similar to that of problem 1, i.e. choose $X^{t+1} = X^t \cup \mbox{argmax}_{j \in V \backslash X^t} f(j | X^t)$. We stop as soon as $f(X^t) = f(V)$, or in other words, we achieve a set which covers all the concepts.

\paragraph{Lazy Greedy Implementations: } Each of the greedy algorithms above admit lazy versions which run much faster than the worst case complexity above~\cite{minoux1978accelerated}. The idea is that instead of recomputing $f(j | X^t), \forall j \notin ^t$, we maintain a priority queue of sorted gains $\rho(j), \forall j \in V$. Initially $\rho(j)$ is set to $f(j), \forall j \in V$. The algorithm selects an element $j \notin X^t$, if $\rho(j) \geq f(j | X^t)$, we add $j$ to $X^t$ (thanks to submodularity). If $\rho(j) \leq f(j | X^t)$, we update $\rho(j)$ to $f(j | X^t)$ and re-sort the priority queue. The complexity of this algorithm is roughly  $O(k n_R T_f)$, where $n_R$ is the average number of re-sorts in each iteration. Note that $n_R \leq n$, while in practice, it is a constant thus offering almost a factor $n$ speedup compared to the simple greedy algorithm.
\paragraph{Further Improvement via Memoization: } One of the parameters in the lazy greedy algorithms is $T_f$, which involves evaluating $f(X \cup j) - f(X)$. One option is to do a na\"{\i}ve implementation of computing $f(X \cup j)$ and then $f(X)$ and take the difference. However, due to the greedy nature of algorithms, we can use memoization and maintain a precompute statistics $p_f(X)$ at a set $X$, using which the gain can be evaluated much more efficiently. At every iteration, we evaluate $f(j | X)$ using $p_f(X)$, which we call $f(j | X, p_f)$.  We then update $p_f(X \cup j)$ after adding element $j$ to $X$. Table 1 provides the precompute statistics, as well as the computational gain for each choice of a submodular function $f$. In particular, it is easy to see that evaluating $f(j | X)$ na\"{\i}vely is much more expensive than evaluating $f(j | X, p_X)$.

\subsection{Instantiations of the Submodular Functions}
Having discussed the choices of the submodular functions, algorithms and the features extracted in the preprocessing, we provide more details on the implementations in our system.

We first investigate the choices of the similarity functions, features and labels. Start with
extractive summarization, obtaining keyframes and skims. For the Facility Location function and the disparity min function, we define the similarity kernel as:
\begin{align*}
s(i, j) = &\langle F_s(i), F_s(j) \rangle + \langle F_o(i), F_o(j) \rangle \\ + &\mbox{corr}(H(i), H(j))
\end{align*}
where $F_s$ represent normalized Deep Scene Features extracted using GoogleNet on Places205~\cite{zhou2014learning}, $F_o$ represents normalized Deep Object features using GoogleNet on ImageNet~\cite{szegedy2015going}, $H$ represents the normalized color histogram features. The similarity function we consider is the sum of the similarities from the normalized scene, object and color features. Since the disparity min function uses a distance function, we use $d_{ij} = 1 - s_{ij}$. For Feature based functions, the feature-set $\mathcal F$ is a concatenation of the scene features $F_s$ and object features $F_o$. In order to define the Set Cover function, we define $U_i$ as the scene and YOLO object labels correesponding to the Image. Recall that the labels for scenes and objects were chosen based on a pre-defined threshold (i.e. select scene and objects labels if the probability for the label is greater than a threshold). The Probabilistic Set Cover function is defined via a concatination of the probabilities from the scene and object models.

Query based summarization for keyframes and snippets is very similar to the extractive summarization, except that we first get a groundset $V_q$ which is related to the query. The queries, are either objects, scenes, faces/humans with age and gender, text in the video, as well as meta data like subtitles etc. In each case, we use the corresponding classifiers as described in Section 3.2 to obtain the set of frames. Once we obtain $V_q$, the summary is obtained in a manner similar to the above.

Finally, for entity or concept based summarization (both in the extractive and query based context), we extract the entities from the videos. Entities we consider are objects, scenes, humans and faces. In the preprocessing stage, we extract all these entities along with their position. For the summarization, we use the features related to the entities.
\begin{enumerate}
\item \textbf{Faces}: In the case of faces, we use the VGG Face model from~\cite{parkhi2015deep}, pretrained on Celeb Face data for Face recognition. For Facility Location and Disparity Min functions, we use a similarity function defined via normalized Face features. Feature based functions use the normalized features, while the set cover function uses the labels from the Face recognition model. Similarly the probailistic set cover function uses the probability from the final layer. The face recognition model provides a diversity in terms of the people (i.e. it attempts to find a diverse set of people). We also use the age and gender models from~\cite{levi2015age}. The Submodular functions are defined very similar to the face recognition model case, except that we combine the age and gender features. For the labels and probability, we obtain a cross product between the labels and probabilities for the age and gender models (i.e. the classes are \emph{Male 0-2 years, Female 0-2 years, Male 2-4 years} and so on. In this case, we attempt diversity in age and gender distributions.

\item \textbf{Scenes: } In the case of scenes, we define features by combining the scene features (defined above) and color histogram features. In other words, for each frame, define $s(i, j) = \langle F_s(i), F_s(j) \rangle + \mbox{corr}(H(i), H(j))$. The Feature based function uses features extracted from the scene model, while for the set cover and the probabilistic set cover, we again use the labels and the probabilities from the scene model. Unlike the diversity and representation functions, the coverage functions defined above do not use the color information, and hence do not distinguish between, say, a bluish ocean and a green ocean scene. Hence, we also classify the color of the scene using the average HSV content of the image. In particular, we classify the frame into 12 different color bins, i.e. Red, Green, Blue, Black, White, Grey, Purple, Violet, Yellow, Orange, Brown and Pink. We take a cross product between the Color class and the scene category, and use those as the labels and probabilities.

\item \textbf{Objects: } The objects are localized using YOLO~\cite{redmon2016you}. We extract features from GoogLeNet~\cite{krizhevsky2012imagenet, szegedy2015going}.
along with color histogram~\cite{swain1991color}. The similarity kernel we use here is $s(i, j) = \langle F_o(i), F_o(j) \rangle + \mbox{corr}(H(i), H(j))$. For the feature based function we use the object features from GoogLeNet~\cite{krizhevsky2012imagenet, szegedy2015going} model pretrained on Imagenet. For the set cover and probabilistic set cover functions, we use the labels and probabilities respectively from YOLO~\cite{redmon2016you}. Rather than just using the object labels and probabilities, we also classify the color of the object in a manner similar to the scene concepts.
\item \textbf{Humans: } Similar to objects, we localize the humans using Human detectors~\cite{dalal2005histograms, felzenszwalb2008discriminatively, redmon2016you}. We use a GoogleNet model pretrained on Imagenet objects as the feature representation~\cite{krizhevsky2012imagenet, szegedy2015going}. We combine this with Color histogram features to compute the similarity kernel, similar to the objects above. Similarly, for the feature based functions, set cover and probabilistic set cover functions, we use the methodology exactly similar to the above.
\end{enumerate}
\begin{figure}

\includegraphics[width=0.45\textwidth]{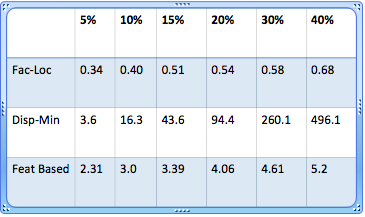}
\caption{Timing results in seconds for summarizing a two hour video for various submodular functions}
\label{timing}
\end{figure}

Next we discuss the choice of the submodular functions. We do not consider DPPs in our system just because they are too computationally expensive (just computing $T_f$ scales as $O(n^3)$). In contrast, Facility Location, Disparity Min and Feature based functions are linear in $n$. The Facility Location and Disparity Min are instantiated using Similarity Kernels discussed above. Feature Based functions are defined directly via features, and we use the deep features as described above. In the case of the Set cover and probabilistic set cover functions, we use the labels and probabilities respectively from the deep models as the concepts. Table 1 provides the complexity of evaluating the gain using the pre-computation statistics. In practice, we observe that the complexity of the greedy algorithm is $O(kT_f)$. The user can select the submodular function to use for summarization.

Finally, we discuss the different types of summarization and the resulting optimization problems. In the case of extractive summarization, which generates a video skim, we support both a fixed length snippet (say 2 seconds or 3 seconds) and a variable length one (determined by shot transitions). The resulting algorithms correspond to the greedy algorithms for Problems 1 and 2. The user here defines the desired length of the summary (in seconds). The user can also choose the coverage option, where he specifies the coverage fraction. The choice of $S(X)$ depends on whether he chooses the fixed or the variable length snippet, and the snippet size. In the case of key frames, we sample the video at a given framerate and the user selects the target size (in terms of the number of images) or the coverage fraction, which then becomes an instance of Problems 1 and 3 respectively. In the case of entity based summarization, the ground set is comprised of the set of entities, depending on the concept chosen (faces, objects, humans or scenes). Again, it becomes an instance of Problem 1 or 3, depending on whether the user wants a summary of a fixed number of entities or a coverage fraction. Note that in this case, the summary consists of entities (for example, cut out objects or faces from the video). Based on the selected objects, we map them back to frames or snippets in the video, also providing the option to view the summary as a video skim or key frame for the chosen concept.

\begin{figure*}
\includegraphics[width=\textwidth]{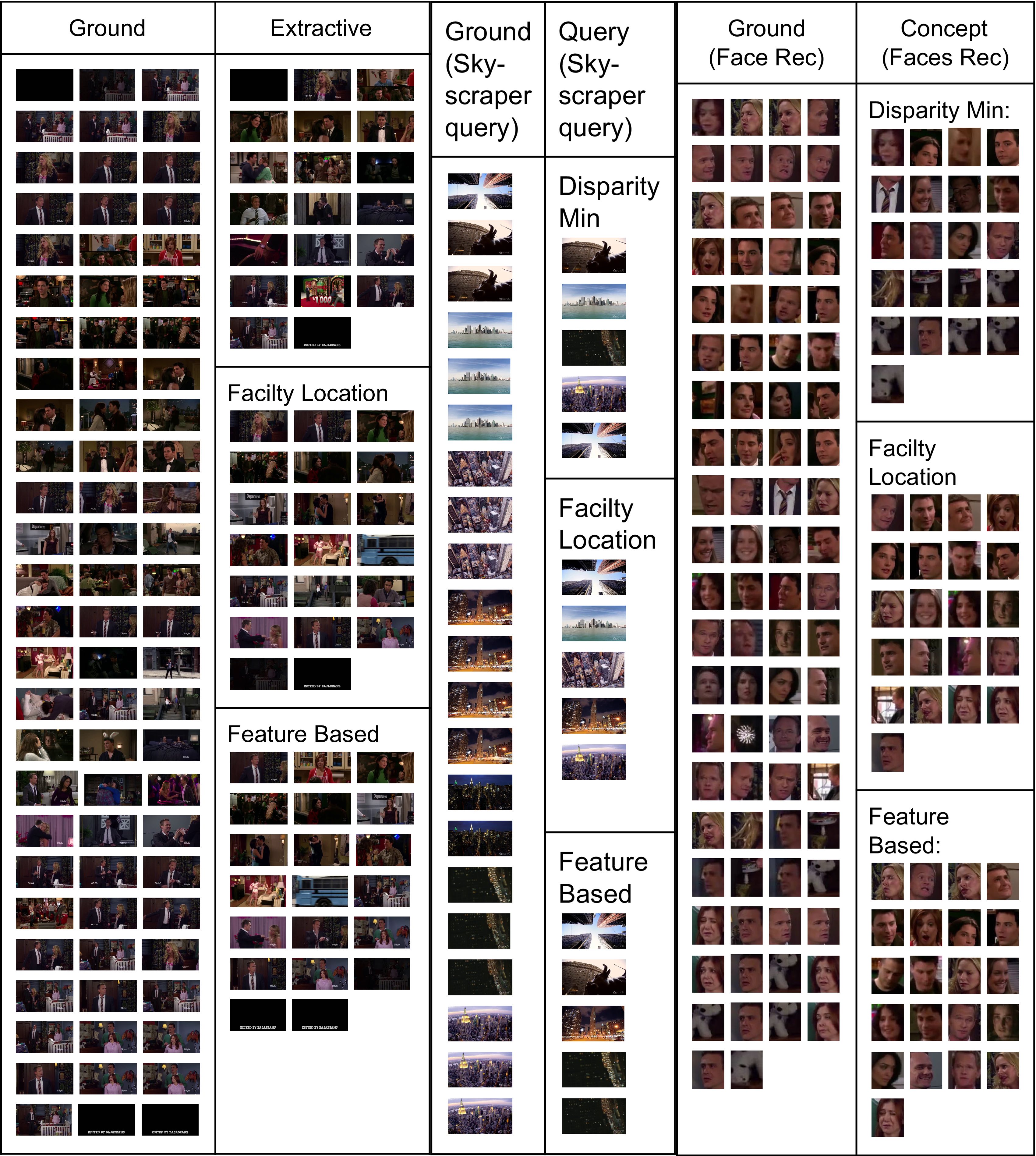}
\caption{Illustration of the Results. The left tab shows the ground set and results for extractive summarization, the middle tab shows the groundset and results for query based summarization and the right tab shows the results for concept based summarization with Faces.}
\label{results}
\end{figure*}

\section{Results}
Our system is implemented in C++. We use Caffe~\cite{jia2014caffe} and DarkNet~\cite{redmon2016you} for deep CNNs and OpenCV~\cite{bradski2008learning} for other computer vision tasks. In the preprocessing step, we first detect and localize all faces using Histogram of Gradients~\cite{dalal2005histograms} and objects and people using Darknet with YOLO~\cite{redmon2016you}. We then extract features using Face VGG (for the faces), GoogLeNet and YOLO for all objects and people, Scene GoogleNet, Color Histograms and Object GoogLeNet for the entire frames. We do this at a processing framerate of one FPS. We store the analysis database in JSON format, and we then run the various summarization modes.

Figure~\ref{results} shows the results for extractive summarization as keyframes, extractive summarization on concepts or entities and query based summarization on keyframes. The concept and extractive summarization are performed on a one minute 21 second video from \emph{How I met your Mother} (Link: \href{https://www.youtube.com/watch?v=1pFSd_MvpBc}{https://www.youtube.com/watch?v=e28t24QE8DA}). We show the results as keyframes, and the ground set is sampled from the video at 1 FPS, resulting in a groundset of 81 frames (shown in the first column of the figure). We then demonstrate the summary obtained via Disparity Min, Facility Location and Feature Based Function (using $\log$ as the concave function). Notice that the Disparity Min function gets a few outliers (like the first frame which is a black frame), while the Facility Location gets representatives from each cluster. Since Disparity Min does not focus on obtaining representatives from each cluster, it misses a few clusters (for example, the cluster of images from 8 to 12), while Facility Location gets representation from each cluster. Finally, it is evident that the Feature based function gets a uniform coverage of all the concepts in the image, and since it does not focus directly on diversity or representation, gets some redundant images (like the last two in the summary). The middle tab shows the results of concept based summarization with faces, again from the same video. Our face detector finds 82 faces, some of which are false positives, and the second column in the middle tab shows the results. Disparity Min gets all the diverse faces, but also some of the outliers, while Facility Location and the Feature Based Function get all the Faces. Finally, the third tab shows the results for query based summarization. For that, we demonstrate the results on a one minute seventeen seconds video from \href{https://www.youtube.com/watch?v=e28t24QE8DA}{https://www.youtube.com/watch?v=e28t24QE8DA}, and we extract query frames related to the query \emph{skyscraper}. We obtain 24 query frames, and choose a summary of five images. The last column shows the results using the Facility Location, Disparity Min and Feature Based Functions.

\begin{figure*}
 \includegraphics[width=\textwidth]{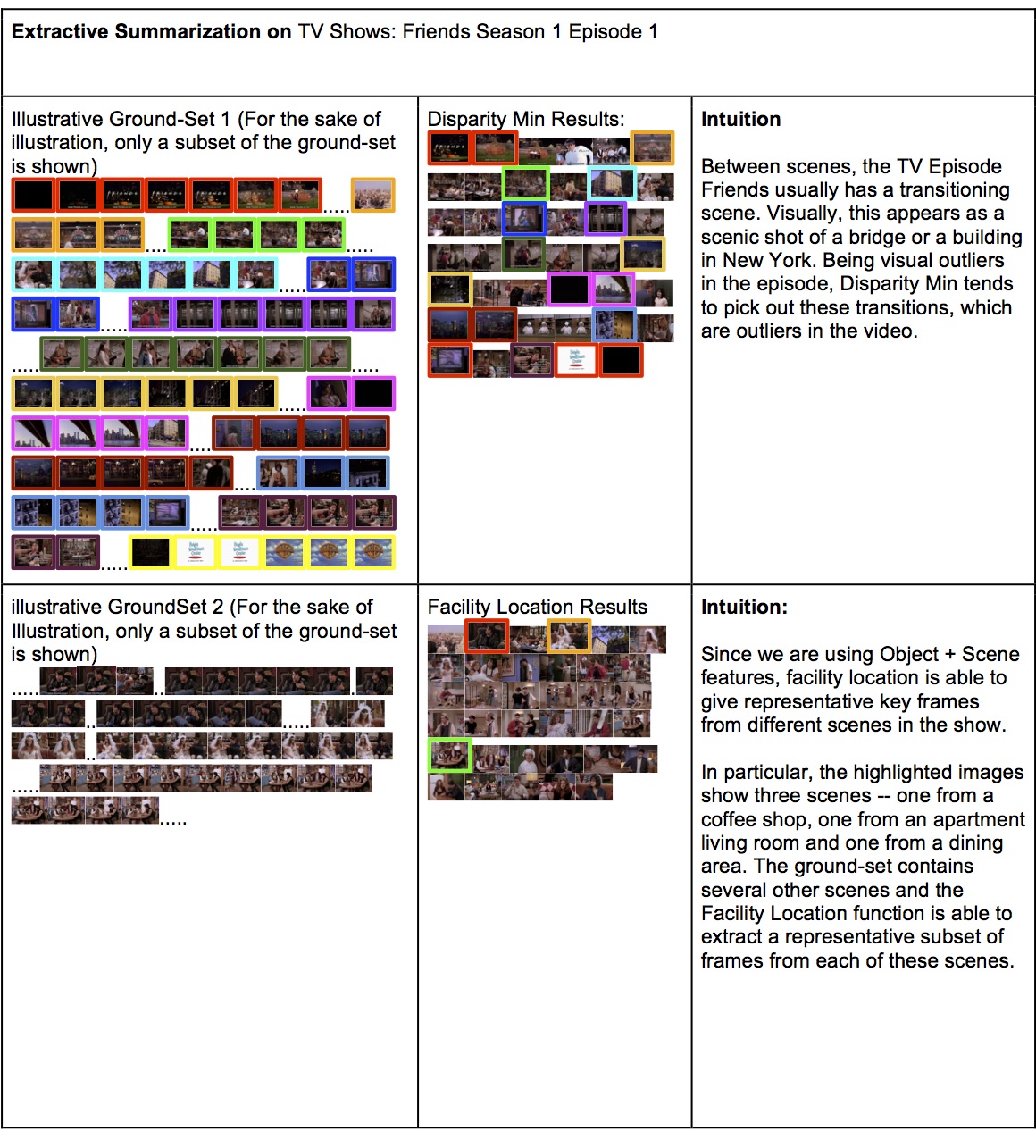}
 \caption{Results on Extractive Summarization on the TV Show, Friends Season 1 Episode 1. Comparing the results from the Disparity and Facility Location functions. See the intuition on the right hand side column.}
 \label{results1}
 \end{figure*}

\begin{figure*}
  \includegraphics[width=\textwidth]{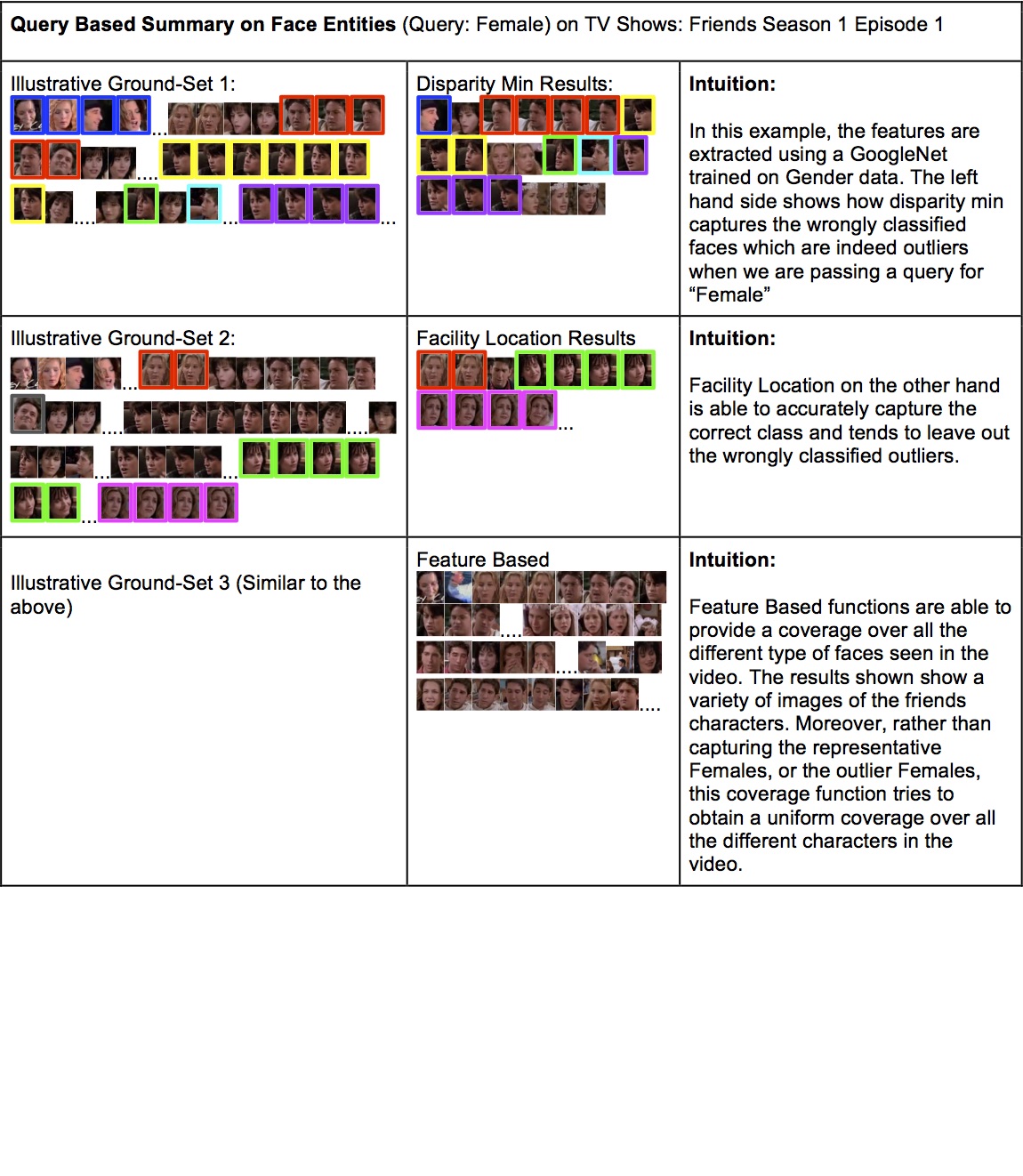}
  \caption{Query Based Summary on Face Entities (Query: Female) on the TV Show Friends Season 1 Episode 1. We compare the results from Disparity, Facility Location and Disparity Functions. See the intuition on the right hand side of the column.}
 \label{results2}
 \end{figure*}

\begin{figure*}
  \includegraphics[width=\textwidth]{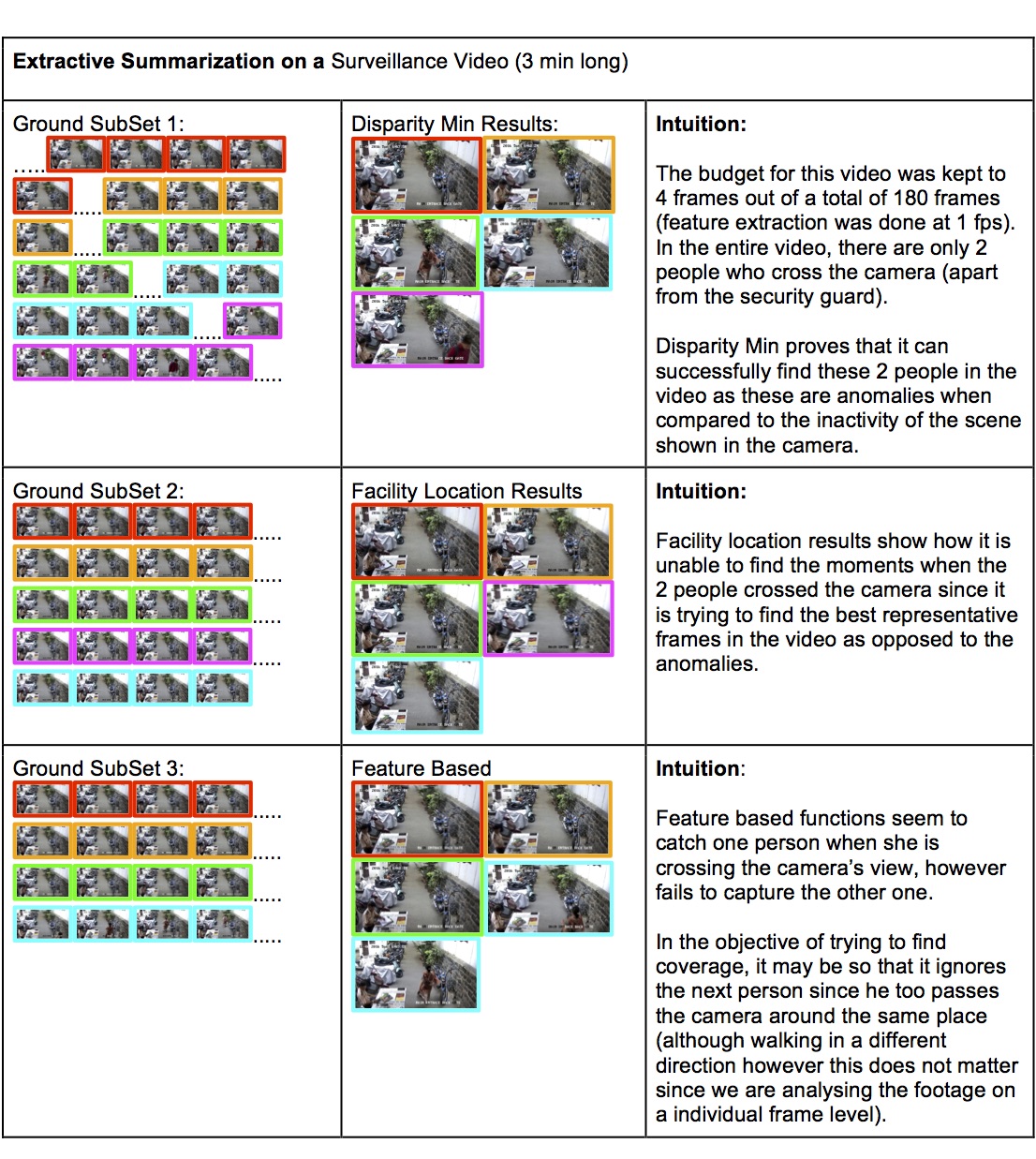}
  \caption{Extractive Summarization Results on Surveillance Video. We compare Disparity function with the Facility Location and the Feature based functions. See the right hand side of the figure for the intuitions.}
 \label{results3}
 \end{figure*}

\begin{figure*}
  \includegraphics[width=\textwidth]{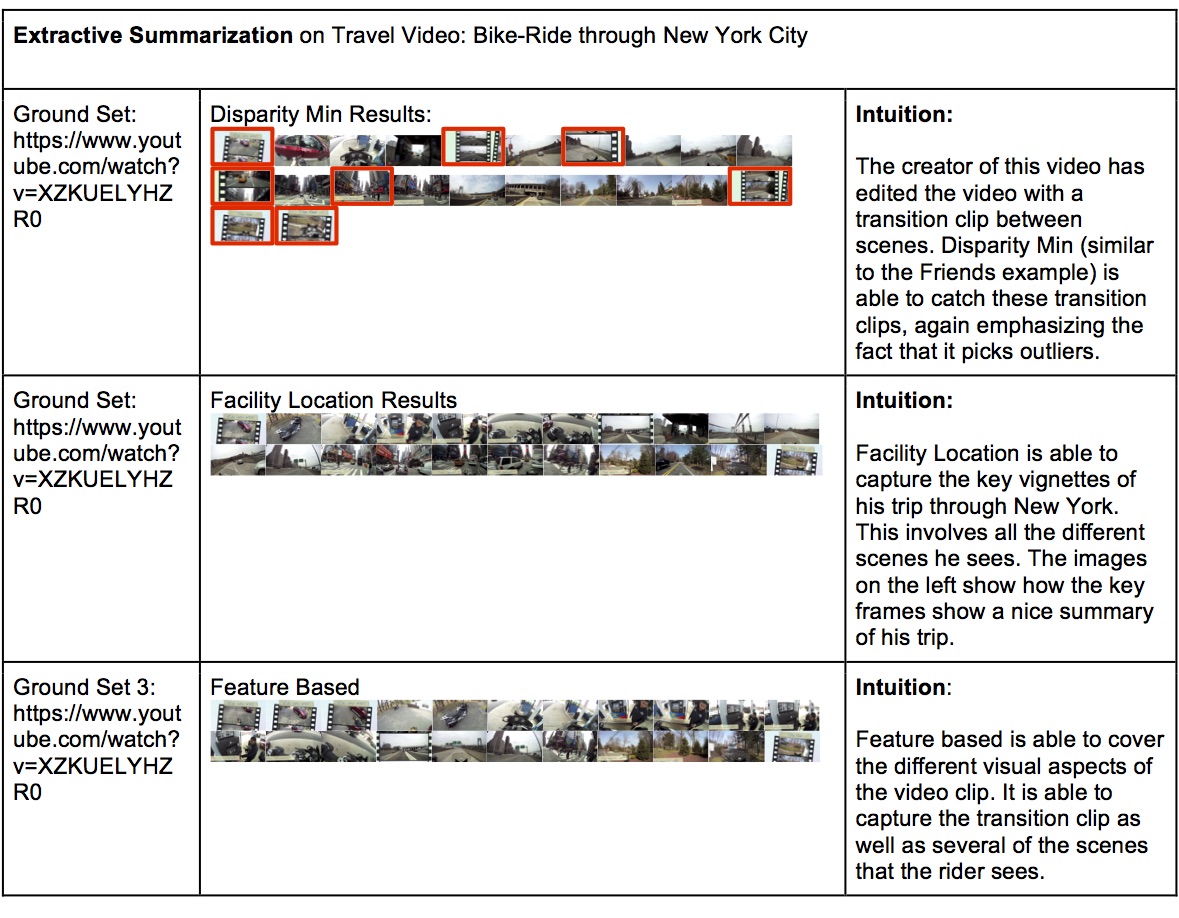}
  \caption{Extractive Summarization Results on a Travel Video. We compare Disparity, Facility Location and the Feature based functions. The intuition is on the right of the figure.}
 \label{results4}
 \end{figure*}

\begin{figure*}
  \includegraphics[width=\textwidth]{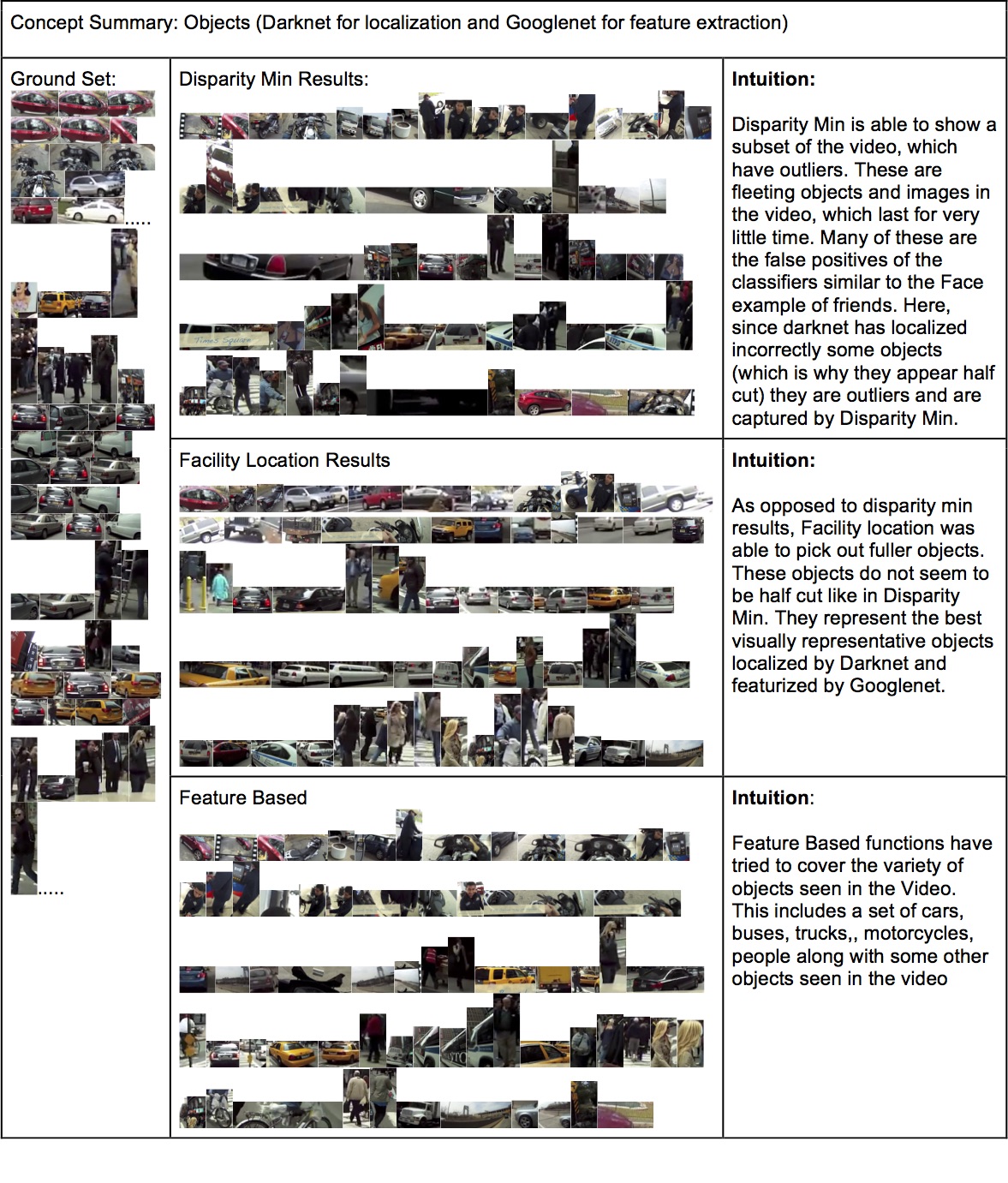}
  \caption{Concept based Summarization on Objects, shown on a Travel Video. We compare Disparity, Facility Location and the Feature based functions. The intuition is on the right of the figure.}
 \label{results5}
 \end{figure*}

Figures~\ref{results1}, \ref{results2}, \ref{results3}, \ref{results4} and \ref{results5} further demonstrate the performance of our system on various settings, including extractive summarization, query based and concept based summarization. Results from Figures~\ref{results1} and \ref{results2} are on a TV show of Friends, while Figure~\ref{results3} demonstrate the results on a Surveillance video. Finally Figures~\ref{results4} and \ref{results5} demonstrate the results on a Travel video. In each case, we explain the intuition from the results. In particular, we show that Disparity Min extracts critical anomalies in video, which is important, for example in surveillance videos, while Facility Location picks the representative frames. The Feature based functions on the other hand, attempt at getting a uniform coverage over the concepts.

Finally, we demonstrate the computational scalability of our framework. Figure~\ref{timing} shows the results of the time taken for Summarization for a two hour video (in seconds). The groundset size is $|V| = 7200$. The Facility Location function obtains the summary within less than a second, while the Feature based functions take a few seconds. Disparity-Min takes a longer time for larger summary sizes. However, the complexity of Disparity Min depends only on the size of the summary, which in this case for a 5 percent size is ten minutes. In general, users like to view summaries which are a few minutes of length (within ten to twenty minutes), in which case the Disparity Min is also scalable. All our experiments were performed on Intel(R) Xeon(R) CPU E5-2603 v3 @1.6 GHz (Dual CPU) with 32 GB RAM. We used a NVIDIA 1070 GTX 8GB GPU for the Deep Learning. For the two hour video, the preprocessing took around 30 minutes on a single GPU. It would be much faster on multiple GPUs and moreover, this is typically done only once.

\section{Acknowledgements}
The authors would like to thank Mick Das and Manoj Joshi for discussions and for painstakingly going through the paper and providing their feedback.
\bibliographystyle{ieee}
\bibliography{iccv}

\end{document}